\documentclass[10pt,twocolumn,letterpaper]{article}

\usepackage[table]{xcolor}

\usepackage[pagenumbers]{iccv} % To force page numbers, e.g. for an arXiv version

\usepackage{fix-cm}
\usepackage{mathrsfs}

\newcommand{\titlespace}{$\quad$}
\newcommand{\titleheight}{\vspace{3pt}}

\definecolor{v}{RGB}{255, 192, 203}
\definecolor{xiang}{RGB}{0, 192, 203}
\definecolor{ranjay}{RGB}{0, 80, 103}
\definecolor{george}{RGB}{0, 212, 255}
\definecolor{NavyBlue}{rgb}{0.0, 0.0, 0.5}

\newcommand{\cfm}{\textsc{$\Delta$FM}\xspace}
\newcommand{\appropto}{\mathrel{\vcenter{
\offinterlineskip\halign{\hfil$##$\cr
\propto\cr\noalign{\kern2pt}\sim\cr\noalign{\kern-2pt}}}}}
\definecolor{iccvblue}{rgb}{0.21,0.49,0.74}
\usepackage[pagebackref,breaklinks,colorlinks,allcolors=iccvblue]{hyperref}

\hypersetup{
    colorlinks,
    linkcolor={red!50!black},
    citecolor={blue!50!black},
    urlcolor={blue!80!black}
}
\usepackage{url}
\usepackage{graphicx, mathtools, amsmath, amssymb, subcaption, multirow, overpic, textpos, pifont, xfrac, bbm, xparse, array}
\usepackage{microtype}
\usepackage[skip=4pt]{caption}
\usepackage{pythonhighlight}
\usepackage{listings}
\usepackage{inconsolata}
\usepackage{epsfig}
\usepackage{booktabs}
\usepackage[scaled=0.8]{FiraMono}
\usepackage[normalem]{ulem}
\usepackage{wrapfig}
\usepackage{adjustbox}
\usepackage{twemojis}
\usepackage{tabularx}
\usepackage{algorithm}
\usepackage{algpseudocode}
\newcommand{\modelname}{Contrastive Flow Matching\xspace}

\def\paperID{6784} % *** Enter the Paper ID here
\def\confName{ICCV}
\def\confYear{2025}

\title{Contrastive Flow Matching}

\author{%
  \centerline{George Stoica$^{1 2\dagger}$\titlespace{}\titlespace{} Vivek Ramanujan$^{2\diamondsuit}$\titlespace{}\titlespace{} Xiang Fan$^{2\diamondsuit}$\titlespace{}\titlespace{}}\\
  \centerline{Ali Farhadi$^2$\titlespace{}\titlespace{}Ranjay Krishna$^2$\titlespace{}\titlespace{}Judy Hoffman$^1$}\titleheight{}\\
    \centerline{$^1$Georgia Tech\titlespace{}$^2$University of Washington}\titleheight{}\\
    \centerline{$^\dagger$Correspondence to: \texttt{gstoica3@gatech.edu} \titlespace{} $^\diamondsuit$Equal Contribution} \\
}
\newlength\savewidth
\newcommand{\tablestyle}[2]{\setlength{\tabcolsep}{#1}\renewcommand{\arraystretch}{#2}\centering\footnotesize}
\renewcommand{\paragraph}[1]{\vspace{1.25mm}\noindent\textbf{#1}}

\usepackage{array} % Ensure array package is included
\newcolumntype{x}[1]{>{\centering\arraybackslash}p{#1pt}}
\newcolumntype{y}[1]{>{\raggedright\arraybackslash}p{#1pt}}
\newcolumntype{z}[1]{>{\raggedleft\arraybackslash}p{#1pt}}
\def\name{\cfm}

\begin{document}

\twocolumn[{
\maketitle
\begin{center}
    \vskip -0.3in
    \captionsetup{type=figure}
    \includegraphics[width=1.\linewidth]{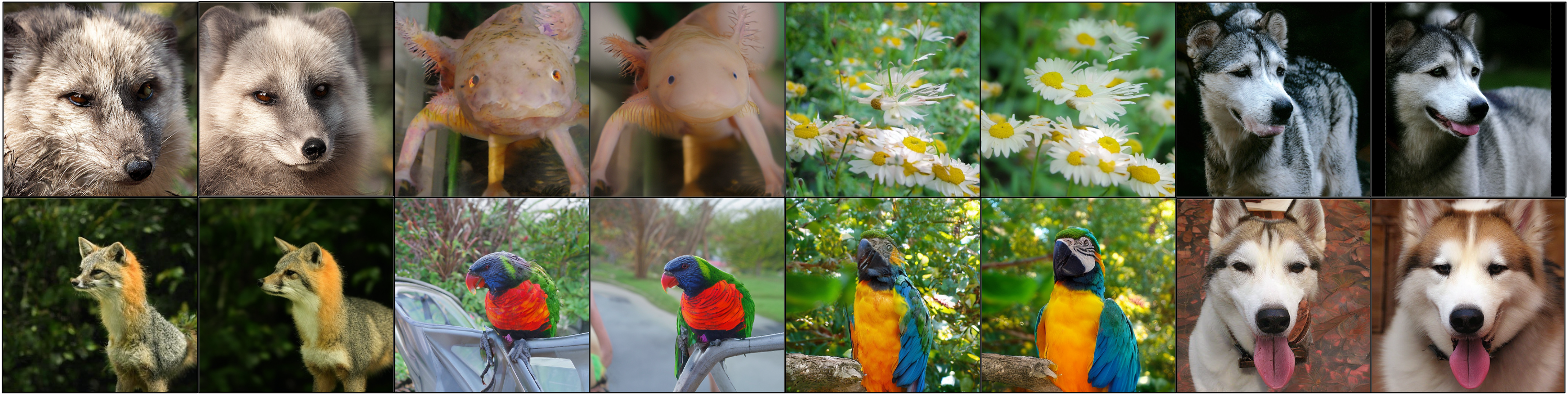}
    \caption{
    {\bf Training with Contrastive Flow-Matching (\cfm) improves natural image generation.} 
    (\textbf{left} is baseline, \textbf{right} is with \cfm)
    Here we show comparisons between images generated by diffusion models trained on ImageNet-1k ($512\times 512$). Each pair of images is generated with the same class and initial noise to ensure similar image structure for comparability. We see that our \cfm objective encourages significantly more coherent images and improves the consistency of global structure.
    }
    % \vspace{4pt}
    \label{fig:teaser}
\end{center}

}]

\begin{abstract}

Unconditional flow-matching trains diffusion models to transport samples from a source distribution to a target distribution by enforcing that the flows between sample pairs are unique.
However, in conditional settings (e.g., class-conditioned models), this uniqueness is no longer guaranteed—flows from different conditions may overlap, leading to more ambiguous generations.
We introduce Contrastive Flow Matching, an extension to the flow matching objective that explicitly enforces uniqueness across all conditional flows, enhancing condition separation. 
Our approach adds a contrastive objective that maximizes dissimilarities between predicted flows from arbitrary sample pairs.
We validate \modelname by conducting extensive experiments across varying model architectures on both class-conditioned (ImageNet-1k) and text-to-image (CC3M) benchmarks.
Notably, we find that training models with Contrastive Flow Matching (1) improves training speed by a factor of up to $9\times$, (2) requires up to $5\times$ fewer de-noising steps and (3) lowers FID by up to $8.9$ compared to training the same models with flow matching.
We release our code at: \url{https://github.com/gstoica27/DeltaFM.git}.
\end{abstract}

\section{Introduction}
\label{sec:intro}

% STATE OF THE WORLD and PROBLEM WITH THE STATE
Flow matching for generative modeling trains continuous normalizing flows by regressing ideal probability flow fields between a base (noise) distribution and the data distribution~\cite{lipman2023flow}. This approach enables straight-line generative trajectories and has demonstrated competitive image synthesis quality. However, for conditional generation (e.g., class-conditional image generation), vanilla flow matching models often produce outputs that resemble an ``average'' of the possible images for a given condition, rather than a distinct mode of that condition. 
In essence, the model may collapse multiple diverse outputs into a single trajectory, yielding samples that lack the expected specificity and diversity for each condition~\cite{ma2024sit, yu2024representation}. 
By contrast, an unconditional flow model---tasked with covering the entire data distribution without any conditioning---implicitly learns more varied flows for different modes of the data. 
Existing conditional flow matching formulations \textit{do not enforce the flows to differ across conditions}, which can lead to this averaging effect and suboptimal generation fidelity.

\begin{figure*}[t]
\centering
    \includegraphics[width=0.95\textwidth]{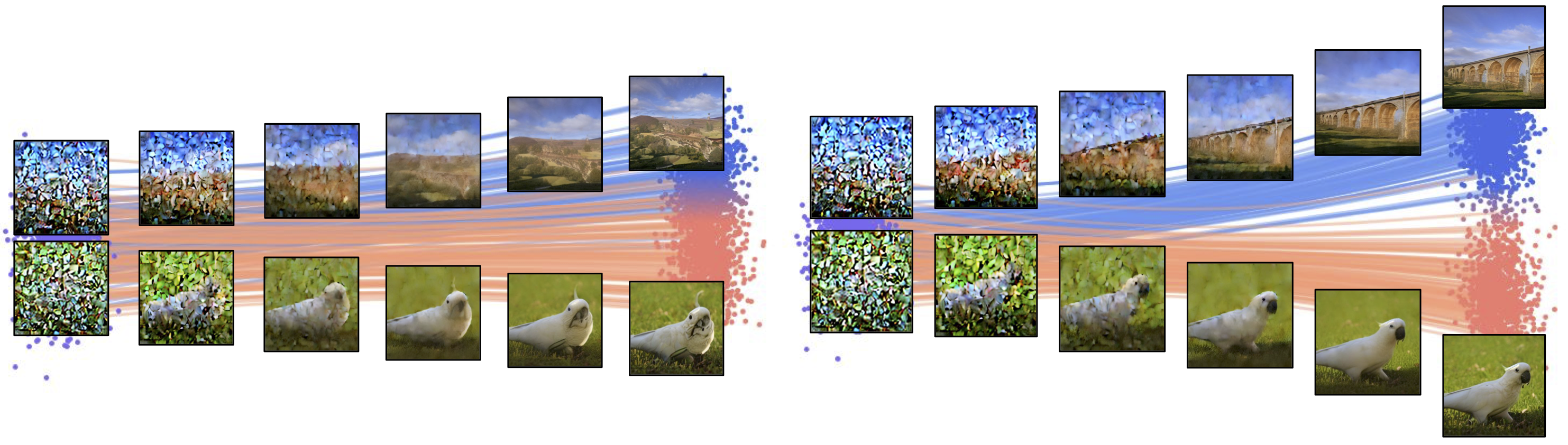}
\caption{
{\bf \cfm{} yields more discriminative and higher quality trajectories.} \textbf{(left)} shows the result of standard flow-matching, where flows are straight but end up overlapping for similar class distributions. \textbf{(right)} shows how the addition of the \cfm objective results in more distinct flows, resulting in images which are more representative of their respective classes.}
\label{fig:flows_expectations}
\end{figure*}

% WHAT HAVE PEOPLE DONE SO FAR.
To address these limitations and improve generation quality, recent work has explored enhancements to structure the generator's representations and also proposed inference-time guidance strategies. For example, one approach is to incorporate a \textit{REPresentation Alignment} (REPA) objective to structure the representations at an intermediate layer with those from a high-quality pretrained vision encoder~\cite{yu2024representation}. By using feature embeddings from a DINO self-supervised vision transformer~\cite{caron2021emerging, oquab2024dinov}, the generative model's hidden states are guided toward semantically meaningful directions. This representational alignment provides an additional learning signal that has been shown to improve both training convergence and final image fidelity, albeit at the cost of requiring an external pretrained encoder and an auxiliary loss term. Another popular technique is \textit{classifier-free guidance} (CFG) for conditional generation~\cite{ho2022classifier}, which involves jointly training the model in unconditional and conditional modes (often by randomly dropping the condition during training). At inference time, CFG performs two forward passes---one with the conditioning input and one without---and then extrapolates between the two outputs to push the sample closer to the conditional target~\cite{ma2024sit, ho2022classifier}. 
While CFG can significantly enhance image detail and adherence to the prompt or class label, it doubles the sampling cost and complicates training by necessitating an implicit unconditional generator alongside the conditional ones~\cite{autoguidance,desai2024improvingimagesynthesisdiffusionnegative,desai2024improvingimagesynthesisdiffusionnegative}.

% OUR ORTHOGONAL CONTRIBUTION.
We propose \textbf{Contrastive Flow Matching} (\cfm), a new approach that augments the flow matching objective with an auxiliary contrastive learning objective. \cfm encourages more diverse and distinct conditional generations. It applies a contrastive loss on the flow vectors (or representations) of samples within each training batch, encouraging the model to produce dissimilar flows for different conditioning inputs. 
Intuitively, this loss penalizes the model if two samples with different conditions yield similar flow dynamics, thereby explicitly discouraging the collapse of multiple conditions onto a single ``average'' generative trajectory. 
As a result, given a particular condition, the model learns to generate a unique flow through latent space that is characteristic of that condition alone, leading to more varied and condition-specific outputs. 
Importantly, this contrastive augmentation is \textit{complementary} to existing methods. 
It can be applied along with REPA, further ensuring that flows not only align with pretrained features but also remain distinct across conditions. Likewise, it is compatible with classifier-free guidance at sampling time, allowing one to combine its benefits with CFG for even stronger conditional signal amplification. 

% TECHNICAL CONTRIBUTION
Inspired by contrastive training objectives, \cfm applies a pairwise loss term between samples in a training batch: for each positive sample from the batch, we randomly sample a negative counterpart. We then encourage the model to not only learn the flow towards the positive sample but also to learn the flow away from the negative sample. This is achieved by adding a contrastive loss to the flow matching objective, which promotes class separability throughout the flow. Our method is simple to implement and can be easily integrated into existing diffusion models without any additional data and with minimal computational overhead.

% EXPERIMENTS, HIGHLIGHTS, ENDING WITH BROADER MESSAGE
We validate the advantages of \cfm through (1) extensive experiments on conditional image generation using ImageNet images across multiple SiT~\cite{ma2024sit} model scales and training frameworks~\cite{ma2024sit, yu2024representation}, and (2) text-to-image experiments on the CC3M~\cite{sharma2018conceptual} with the MMDiT~\cite{esser2024scaling} architecture. Thanks to contrastive flows, \cfm consistently outperforms traditional diffusion flow matching in quality and diversity metrics, achieving up to an $8.9$-point reduction in FID-50K on ImageNet, and $5$-point reduction in FID on the whole CC3M validation set. It is also compatible with recent significant improvements in the diffusion objective, such as Representation Alignment (REPA)~\cite{yu2024representation}. By encouraging class separability, \cfm is able to efficiently reach a given image quality with \textbf{$5\times$ fewer} sampling steps than a baseline Flow Matching model, translating directly to faster generation. It also enhances training efficiency by up to \textbf{9$\times$}. 
Finally, \cfm stacks with classifier-free guidance, lowering FID by {\bf 5.7\%} compared to flow matching models.

\section{Related works}
\label{sec:related_works}

Our work lies in the domain of image generative models, primarily diffusion and flow matching models. We augment flow-matching with a contrastive learning objective to provide an alternative solution to classifier free guidance.

Generative modeling has rapidly advanced through two primary paradigms: diffusion-based methods~\cite{Ho2020, Song2021} and flow matching~\cite{lipman2023flow}. 
\textbf{Denoising diffusion models} typically rely on stochastic differential equations (SDEs) and score-based learning to iteratively add and remove noise~\cite{Ho2020}. Denoising diffusion implicit models (DDIMs) \cite{Song2021} reduce this sampling complexity by removing non-determinism in the reverse process, while progressive distillation \cite{Salimans2022} further accelerates inference by shortening the denoising chain. Advanced ODE solvers \cite{Chen2018} and distillation methods \cite{vahdat2021scorebasedgenerativemodelinglatent} have also enhanced sampling efficiency.
Despite their success, diffusion models can be slow at inference due to iterative denoising \cite{Ho2020}. 

\textbf{Flow matching}~\cite{Chen2018} has been designed to reduce inference steps. It directly parameterizes continuous-time transport dynamics for more efficient sampling.
Probability flow ODEs~\cite{Song2021, lipman2023flow} learn an explicit transport map between data and latent distributions. Unlike diffusion models, it bypasses separate score estimation and stochastic noise, which reduces function evaluations and tends to improve training convergence \cite{Chen2018}.
A common type of flow matching algorithm popularized recently is the rectified flow~\cite{liu2022flowstraightfastlearning}, which refines probability flow ODEs through direct optimal transport learning, improving numerical stability and sampling speed. This approach mitigates the high computational burden of diffusion sampling while maintaining high-fidelity image generation with fewer integration steps.

Since both diffusion and flow matching models are trained to match the target distribution of real images, they often produce `averaged' samples that lack the sharp details and strong conditional fidelity~\cite{cfg}. Regardless of how much these models speed up, they often need to be invoked multiple times with unique seed noise to find a high-fidelity sample.
In response, 
\textbf{guidance techniques} have been introduced to substantially promote high-fidelity synthesis. Classifier guidance~\cite{adm}, classifier-free guidance~\cite{cfg}, energy guidance~\cite{chung2022diffusion,zhao2022egsde,cep,song2023loss}, and more advanced methods~\cite{kynkaanniemi2024applying,autoguidance,cfg++,koulischer2024dynamic,shenoy2024gradient} improve fidelity and controllability, without requiring multiple invocations. 
Although they achieve remarkable performance, they typically still require additional computational overhead. CFG requires calling sampling from a second `unconditional' generation and guiding the `conditional' generation away from the unconditional variant~\cite{lcm,dmd,dmd2,sid-lsg}.
We adapt the flow matching objective with a contrastive loss between the transport vectors within a batch. By doing so, we achieve the same benefits of CFG, without the additional overhead of needing to train an unconditional generator or using one during inference.

Contrastive learning was originally proposed for face recognition~\cite{Schroff_2015}, where it was designed to encourage a margin between positive and negative face pairs. In generative adversarial networks (GANs), it has been applied to improve sample quality by structuring latent representations~\cite{cao2017tripletgantraininggenerativemodel}. However, to the best of our knowledge, it has not been explored in the context of visual diffusion or flow matching models. We incorporate this contrastive objective to demonstrate its utility in speeding up training and inference of flow-based generative models.

\section{Background and motivation}
\label{sec:background}
% \textcolor{red}{[NEEDS REVISION]}
We focus on flow matching models~\citep{lipman2023flow} due to its rising popularity as an effective training paradigm for generative models~\cite{lipman2022flow, albergo2022building, albergo2023stochastic}.
In this section, we provide a brief overview of flow matching through the perspective of stochastic interpolants~\cite{albergo2023stochastic, ma2024sit}, as it pertains to our work. 

\paragraph{Preliminaries.} Let $p(x)$ be an arbitrary distribution defined on the reals, and let $\mathcal{N}(0, \mathrm{I})$ be a Gaussian noise distribution.
The objective of flow matching is to learn a transport between the two distributions. 
That is, given an arbitrary $\epsilon\sim \mathcal{N}(0, \mathrm{I})$, a flow matching model gradually transforms $\epsilon$ over time into an $\hat{x}$ that is part of $p(x)$. 
Stochastic interpolants~\cite{albergo2023stochastic, ma2024sit} define this transformation as a time-dependent stochastic process, where transformation steps are summarized as follows,
\begin{align}
    \hat{x}_t = \alpha_t \hat{x} + \sigma_t \epsilon\label{eq:xt_def}
\end{align}
where $\alpha_t$ and $\sigma_t$ are decreasing and increasing time-dependent functions respectively defined on $t\in [0, T]$, such that $\alpha_T = \sigma_0 = 1$ and $\alpha_0 = \sigma_T = 0$.
While theoretically, $\alpha_t, \sigma_t$ need not be linear, linear complexity is often sufficient to obtain strong diffusion models~\cite{ma2024sit, lipman2023flow, yu2024representation}.

\paragraph{Flow matching.} 
Given such a process, flow matching models learn to transport between noise to $p(x)$ by estimating a velocity field over an probability flow ordinary differential equation (PF ODE), $dx_t=v(x_t,t)dt$, whose distribution at time $t$ is the marginal $p_t(x)$. 
This velocity is given by the expectations of $\hat{x}$ and $\epsilon$ conditioned on $x_t$, 
\begin{equation}
    v(x_t, t) = \dot{\alpha}_t\mathbb{E}[\hat{x} | x_t = x] + \dot{\sigma}_t\mathbb{E}[\epsilon | x_t = x],
\end{equation}
where $\dot{\alpha}_t, \dot{\sigma}_t$ are the time-based derivatives of $\alpha_t$ and $\sigma_t$ respectively.
Since, $\hat{x}$ and $\epsilon$ are arbitrary samples from their respective distributions, $v(x_t, t)$ is expected ``direction'' of all transport paths between noise and $p(x)$ that pass through $x_t$ at $t$. 
While the optimal $v(x_t,t)$ is intractable, it can be approximated with a flow-model $v_\theta(x_t, t)$, by minimizing the training objective:
\begin{equation}
    \mathcal{L}^{(FM)}(\theta) = \mathbb{E}\left[ || v_\theta(x_t, t) - (\dot{\alpha}_t\hat{x} + \dot{\sigma}_t\epsilon) ||^2 \right]\label{eq:fm}
\end{equation}
Key to understanding the properties of flow matching is the concept of flow uniqueness~\cite{lipman2023flow}.
That is, flows following the well-defined ODE cannot intersect at \textit{any} time $t\in [0, T)$.
As such, flow models can iteratively refine unique-discriminative features relevant to any $x\sim p(x)$ in each $x_t$, leading to more efficient and accurate diffusion paths compared to other training paradigms~\cite{lipman2023flow}.

\paragraph{Conditional flow matching.}
Commonly, $p(x)$ may be a marginal distribution over several class-conditional distributions (e.g., the classes of ImageNet~\cite{russakovsky2015imagenet}).
Training models in such cases is nearly identical to standard flow-matching, except that flows are further conditioned on the target distribution class: 
\begin{equation}
    \mathcal{L}^{(FM)}_{cond}(\theta) = \mathbb{E}\left[ || v_\theta(x_t, t, y) - (\dot{\alpha}_t\hat{x} + \dot{\sigma}_t\epsilon) ||^2 \right]\label{eq:cfm},
\end{equation}
where $\hat{x}\sim p(x | y)$. 
Resultant models have the desirable trait of being more controllable: their generated outputs can be tailored to their respective input conditions.
However, this comes at the notable cost of flow-uniqueness.
Specifically these models only generate unique flows compared to others \textit{within} the same class-condition, not necessarily \textit{across} classes.
This inhibits $x_t$'s from storing important class-specific features and leads to poorer quality generations. 
Second, the conditional flow matching objective trains models without knowledge of the distributional spread from other class-conditions, leading to flows that may generate ambiguous outputs when conditional distributions overlap 
. This increases the likelihood of ambiguous generations that form a mixture between different conditions, restricting model capabilities. We study these effects in \Cref{sec:experiments}.

\section{Contrastive Flow-Matching}
\label{sec:method}
We introduce Contrastive Flow Matching (\cfm), a novel approach designed to address the challenges of learning efficient class-distinct flow representations in conditional generative models. 
Standard conditional flow matching (FM) models tend to produce flow trajectories that align across different samples, leading to reduced class separability. 
\cfm extends the FM objective by incorporating a contrastive regularization term, which explicitly discourages alignment between the learned flow trajectories of distinct samples.

\paragraph{Ingredients.} 
Let $\tilde{x} \sim p(x | \tilde{y})$ denote a sample drawn from the data distribution conditioned on an arbitrary class $\tilde{y}$, and let $\tilde{\epsilon} \sim \mathcal{N}(0, I)$ represent an independent noise sample. 
To ensure that the contrastive objective captures distinct flow trajectories, we impose the conditions $\tilde{x} \neq \hat{x}$ and $\tilde{\epsilon} \neq \epsilon$, where $\tilde{y}$ may or may not be equal to $y$. 
Importantly, we do not assume the existence of a time step $t \in [0, T]$ such that $x_t = \alpha_t \tilde{x} + \sigma_t \tilde{\epsilon}$. 
Consequently, $\tilde{x}$ and $\tilde{\epsilon}$ represent truly independent flow trajectories in comparison to $\hat{x}$ and $\epsilon$.

\paragraph{The contrastive regularization.} Given $v_\theta(x_t, t, y)$ and an arbitrary $\tilde{x}, \tilde{\epsilon}$ sample pair, the contrastive objective aims to \textit{maximize} the dissimilarity between the estimated flow of $v_\theta(x_t, t, y)$ from $\epsilon$ to $\hat{x}$, and the independent flow produced by $\tilde{x}, \tilde{\epsilon}$. 
We achieve this by maximizing the quantity,
\begin{equation}
    E\left[ ||v_\theta(x_t, t, y) - (\dot{\alpha}_t \tilde{x} + \dot{\sigma}_t \tilde{\epsilon}) ||^2 \right].\label{eq:contrastive}
\end{equation}
Since $\tilde{x}$ is drawn from the marginal $p(x)$ rather than $p(x|y)$, Equation~\ref{eq:contrastive} trains flow matching models to produce flows that are \textit{unconditionally} unique. 

\paragraph{Putting it all together.} We now define contrastive flow matching as follows,
\begin{equation}
\begin{split}
    \mathcal{L}^{(\cfm)}(\theta) = \mathrm{E}\left[
    \begin{aligned}
        & ||v_\theta(x_t, t, y) - (\dot{\alpha}_t \hat{x} + \dot{\sigma}_t \epsilon) ||^2 \\
        & -\lambda ||v_\theta(x_t, t, y) - (\dot{\alpha}_t \tilde{x} + \dot{\sigma}_t \tilde{\epsilon}) ||^2
    \end{aligned}
    \right]
\end{split}\label{eq:cf}
\end{equation}
where $\lambda \in [0, 1)$ is a fixed hyperparameter that controls the strength of the contrastive regularization.
Thus, \cfm simultaneously encourages flow matching models to estimate effective transports from noise to corresponding class-conditional distributions (the flow matching objective), while enforcing each to be discriminative \textit{across} classes (contrastive regularization). 
Note that \cfm can be thought of as a generalization of flow matching, as \cfm reduces to FM when $\lambda = 0$. We study the effects of varying $\lambda$ in Section~\ref{sec:analysis}.

\paragraph{Implementation.}
Contrastive flow matching (\cfm) is easily integrated into any flow matching training loop, with minimal overhead. 
Algorithm~\ref{alg:cfm} illustrates the implementation of an arbitrary batch step, where \textcolor{NavyBlue}{navy text} marks additions to the standard flow matching objective.
Thus, \cfm solely depends on the information already available to the flow matching objective at each batch step, without computing any additional forward steps.
Furthermore, \cfm seamlessly folds into flow matching training regimes, making it a ``plug-and-play'' objective for existing setups.
\begin{algorithm}
\caption{Contrastive Flow-Matching Batch Step}
\label{alg:cfm}
\begin{algorithmic}[1] % The [1] enables line numbering
\State \textbf{Input:} A model $v_\theta$, batch of $N$ flow examples $F=\{(x_1,y_1,\epsilon_1), \ldots, (x_N,y_N,\epsilon_N)\}$ where $(x_i, y_i)\sim p(x, y)$ and $\epsilon_i\sim \mathcal{N}(0,\mathrm{I})$, $\beta$ learning rate, \textcolor{NavyBlue}{$\lambda=0.05$}.
\State \textbf{Output:} Updated model parameters $\theta$
\State $L(\theta) = 0$
\For{$i$ in range($N$)}
    \State $t\sim U(0,1), x_t = \alpha_t x_i + \sigma_t \epsilon_i$
    \State \textcolor{NavyBlue}{sample $(\tilde{x}, \tilde{y}, \tilde{\epsilon})\sim F, \text{ s.t. } (\tilde{x}, \tilde{y}, \tilde{\epsilon}) \neq (x_i,y_i,\epsilon_i)$}
    \State $\hat{v}=v(x_t, t, y_i), v=\dot{\alpha}_t x_i + \dot{\sigma}_t \epsilon, \textcolor{NavyBlue}{\tilde{v}=\dot{\alpha}_t \tilde{x} + \dot{\sigma}_t \tilde{\epsilon}}$
    \State $L(\theta) += || \hat{v} - v ||^2 - \textcolor{NavyBlue}{\lambda ||\hat{v} - \tilde{v} ||^2}$
    \EndFor
\State $\theta \leftarrow \theta - \frac{\beta}{N}\nabla_\theta L(\theta)$
    % \State Return result
\end{algorithmic}
\end{algorithm}

% \begin{figure}[t]  % 't' places it at the top of the column
%     \centering
%     \begin{subfigure}[b]{\linewidth}
%         \includegraphics[width=\linewidth]{figures/imgs/fm.png}
%         \caption{Models trained with flow-Matching generate distribution mixtures.}
%     \end{subfigure}
%     \vspace{0.2cm}
%     \begin{subfigure}[b]{\linewidth}
%         \includegraphics[width=\linewidth]{figures/imgs/cfg.png}
%         \caption{These generations can be remedied through the use of classifier-free guidance, at the cost of extensive hyperparameter tuning and expensive sampling procedures.}
%     \end{subfigure}
%     \vspace{0.2cm}
%     \begin{subfigure}[b]{\linewidth}
%         \includegraphics[width=\linewidth]{figures/imgs/contrastive_flows.png}
%         \caption{Models trained with contrastive-flows naturally generate discriminative representations across distributions.}
%     \end{subfigure}
%     \caption{{
%     \bf Contrastive flow enables discriminative generations without any inference hyperparameter tuning}. 
%     Figures show the denoising paths of a conditional flow-model trained to diffuse from a common source distribution (purple) to two separate class distributions in blue and red.
%     }
%     \label{fig:flows}
% \end{figure}

\begin{figure}[!ht]
\centering
\begin{center}
    \includegraphics[width=0.95\linewidth]{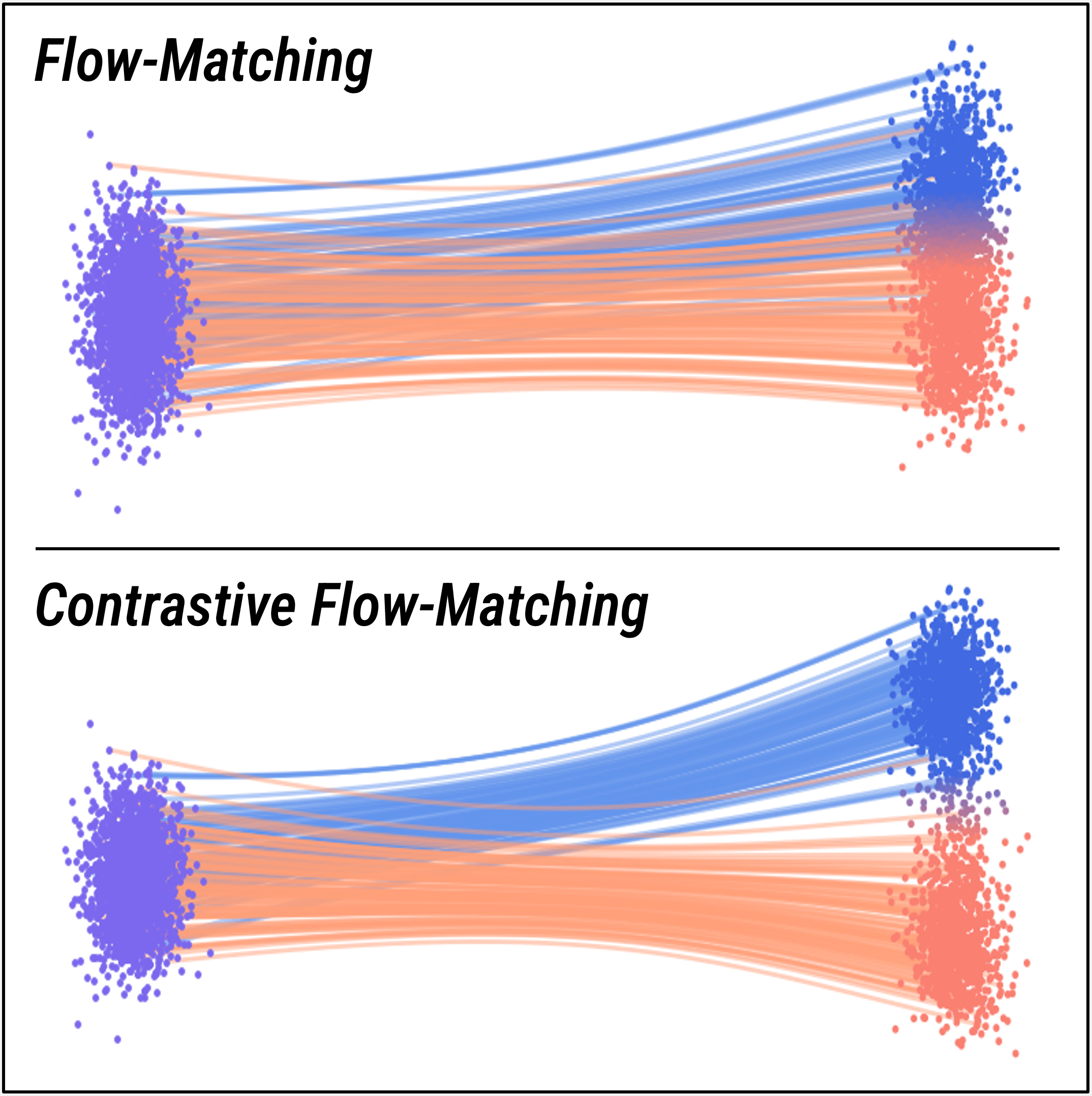}
\end{center}
\caption{
{\bf Contrastive Flow-Matching intrinsically separates flows between classes.} 
We train a small three layer MLP flow-matching model to transport between a two dimensional multivariate noise distribution (\textcolor{violet}{violet}) and two independent  \textcolor{blue}{blue} and \textcolor{orange}{orange} class distributions respectively.
The class distributions are designed to have $\sim 50\%$ overlap, and we plot the learned class-conditioned flows between noise samples and each respective class distribution using class colors.
Top: Flow-matching models learn overlapping transports between distributions, generating outputs that lie in ambiguous regions between the two classes.
Bottom: Contrastive flow-matching models have significantly more discriminative flows, generating class-coherent samples while reducing ambiguity.
}
\label{fig:comparing_flows}
\end{figure}

\paragraph{Discussion.}
Figure~\ref{fig:comparing_flows} illustrates the effects of contrastive flow matching compared to flow matching.
The figure shows the resultant flows after training a small diffusion model in a simple toy-setting.
Specifically, we create a two-dimensional \textcolor{violet}{violet} gaussian noise distribution and two independent two-dimensional class distributions (in \textcolor{blue}{blue} and \textcolor{orange}{orange} respectively) such that the latter distributions have $\approx 50\%$ overlap.
Samples from each distribution are represented as ``dots'', with those in the target distributions colored according to the gaussian \textit{kernel-density estimate} between samples from each class in their respective region. 
We observe that training the model with flow matching (top) create flows with large degrees of overlap between classes, generating samples with lower class-distinction.
In contrast, training the same model with contrastive flow matching (bottom) yields trajectories that are significantly more diverse across classes, while also generating samples which capture distinct features of each respective class.
% \section{Understanding Contrastive Flow-Matching}
% \label{sec:analysis}
% \input{figures/trajectories}
\section{Experiments}
\label{sec:experiments}
We validate contrastive flow-matching (\cfm) through extensive experiments across various model, training and benchmark configurations.
Overall, models trained with \cfm consistently outperform flow-matching (FM) models across \textit{all} settings.

\paragraph{Datasets.}
We conduct both class-conditioned and text-to-image experiments.
% \george{We conduct most of our experiments using the}
We use ImageNet-1k~\cite{imagenet} processed at both ($256\times 256$) and ($512\times 512$) resolutions for our class-conditioned experiments, and follow the data preprocessing procedure of ADM~\cite{adm}  
We then follow~\cite{yu2024representation} and encode each image using the Stable Diffusion VAE~\cite{ldm} into a tensor $z\in\mathbb{R}^{32\times 32\times 4}$. 
For text-to-image (t2i), we use the Conceptual Captions 3M (CC3M) dataset~\cite{sharma2018conceptual} processed at ($256\times 256$) resolution and follow the data processing procedure of~\cite{bao2022all}.
We train all models by strictly following the setup in \cite{yu2024representation}, and use a batch size of 256 unless otherwise specified.
We do not alter the training conditions to be favorable to \cfm, and we always set $\lambda=0.05$ when applicable.

\paragraph{Measurements.} 
We report five quantitative metrics throughout our experiments. 
We report Fréchet inception distance (FID)~\cite{fid}, inception score (IS)~\cite{inception_score}, sFID~\cite{sfid}, precision (Prec.) and recall (Rec.)~\cite{kynkaanniemi2019improved} using 50,000 samples for our class-conditioned experiments.
Similarly, we report FID over the whole validation set in the text-to-image setting.
% we report FID using all the samples in the validation
We use the SDE Euler-Maruyama sampler with $w_t=\sigma_t$ for all experiments, and set the number of function evaluations (NFE) to 50 unless otherwise specified.

\subsection{Contrastive Flow-Matching Improves SiT}\label{sec:sit_vanilla}
\paragraph{Implementation details.}
We train on the state-of-the-art SiT~\cite{ma2024sit} model architecture, using both SiT-B/2 and SiT-XL/2.
\begin{table}[h]
    \centering
    \begin{subtable}{\linewidth}
        \centering
        \tablestyle{5pt}{1.1}
        \begin{tabular}{z{60}x{22}x{20}x{22}x{23}x{22}} 
        \toprule 
        \multirow{2}{*}{ } & \multicolumn{5}{c}{Metrics} \\ 
        \cmidrule{2-6}                  Model & FID $\downarrow$ & IS $\uparrow$ & sFID $\downarrow$ & Prec. $\uparrow$ & Rec. $\uparrow$ \\ 
        \midrule 
        SiT-B/2                         & 42.28   & 38.04   & 11.35 & 0.5     & 0.62 \\ 
        \rowcolor{gray!25}
        % (\cfm) SiT-B/2               & {\bf 33.39}   & {\bf 43.44}   & {\bf 5.67} & {\bf 0.53}     & {\bf 0.63} \\ 
        + Using \name{}               & {\bf 33.39}   & {\bf 43.44}   & \phantom{0}{\bf 5.67} & {\bf 0.53}     & {\bf 0.63} \\ 
        \midrule
        SiT-XL/2                        & 20.01 & 74.15 & 8.45 & 0.63 & {\bf 0.63} \\
        \rowcolor{gray!25}
        % (\cfm) SiT-XL/2              & {\bf 16.32} & {\bf 78.07} & {\bf 5.08} & {\bf 0.66 } & {\bf 0.63} \\ 
        + Using \name{}              & {\bf 16.32} & {\bf 78.07} & {\bf 5.08} & {\bf 0.66 } & {\bf 0.63} \\ 
        \bottomrule \end{tabular}
        \caption{
        {\bf ImageNet-1k (256x256) Results.} \cfm significantly outperforms flow-matching models across nearly all metrics, and matches Recall on SiT-XL/2. 
        }
        \label{tab:sit_imnet256}
    \end{subtable}

    \vspace{1em} % Space between the tables

    \begin{subtable}{\linewidth}
        \centering
        \tablestyle{5pt}{1.1}
        \begin{tabular}{z{60}x{22}x{20}x{22}x{23}x{22}} 
        \toprule 
        \multirow{2}{*}{ } & \multicolumn{5}{c}{Metrics} \\ 
        \cmidrule{2-6}        Model & FID$\downarrow$ & IS$\uparrow$ & sFID$\downarrow$ & Prec.$\uparrow$ & Rec.$\uparrow$ \\ 
        \midrule 
        SiT-B/2               & 50.26 & 33.58 & 14.88 & 0.57 & 0.61 \\ 
        \rowcolor{gray!25}
        % (\cfm) SiT-B/2     & {\bf 41.59} & {\bf 38.20} & {\bf 6.13} & {\bf 0.62} & {\bf 0.63} \\ 
        + Using \name{}     & {\bf 41.59} & {\bf 38.20} & \phantom{0}{\bf 6.13} & {\bf 0.62} & {\bf 0.63} \\ 
        % \cmidrule{1-5}
        \midrule
        SiT-XL/2              & 22.98 & 70.14 & 10.71 & 0.73 & {\bf 0.60} \\ 
        \rowcolor{gray!25}
        % (\cfm) SiT-XL/2    & {\bf 19.67} & {\bf 72.58} & {\bf 4.98} & {\bf 0.76} & {\bf 0.60} \\ 
        + Using \name{}    & {\bf 19.67} & {\bf 72.58} & \phantom{0}{\bf 4.98} & {\bf 0.76} & {\bf 0.60} \\ 
        \bottomrule 
        \end{tabular}
        \caption{
        {\bf ImageNet-1k (512x512) Results.} 
        Models trained with \cfm either substantially outperform or match their flow-matching counterparts in all metrics.
        }
        \label{tab:sit_imnet512}
    \end{subtable}

    \caption{
    SiT~\cite{ma2024sit} results on ImageNet-1k ($256\times 256$; a) and ($512\times 512$; b).
    We train all models for 400K iterations following \citep{yu2024representation}.
    All metrics are measured with the SDE Euler-Maruyama sampler with NFE=50 and without classifier guidance.
    We use $\lambda=0.05$ for all models trained with \cfm and do not change any other hyperparameters.
    $\uparrow$ indicates that higher values are better, with $\downarrow$ denoting the opposite.
    }
    \label{tab:sit_results}
\end{table}

\paragraph{Results.} Table~\ref{tab:sit_results} summarizes our results.
Overall, \cfm dramatically improves over flow-matching in nearly all metrics (only matching the flow-matching SiT-XL/2 model in recall).
Notably, employing \cfm with SiT-B/2 lowers FID by over 8 compared to flow-matching at both ImageNet resolutions, highlighting the strength of \cfm in smaller model scales.
Similarly, \cfm is robust to larger model scales and outperforms FM by over 3.2 FID when using SiT-XL/2.

\subsection{REPA is complementary}
\label{sec:repa}
REPresentation Alignment (REPA)~\cite{yu2024representation} is a recently introduced training framework that rapidly improves diffusion model performance by strengthening its intermediate representations.
Specifically, REPA distills the encodings of foundation vision encoders (e.g., DiNOv2~\cite{caron2021emerging}) into the hidden states of diffusion models through the use of an auxilliary objective. 
Notably, REPA can improve the training speed of vanilla SiT models by over $17.5\times$, while further improving their performances~\cite{yu2024representation}.
\cfm is easily integrated into REPA and only requires replacing the flow-matching objective.

\paragraph{Implementation details.}
We apply REPA on the same SiT models as in Section~\ref{sec:sit_vanilla}, and use the distillation process defined by~\cite{yu2024representation} exactly.
Specifically, we use distill DiNOv2~\cite{caron2021emerging} ViT-B~\cite{dosovitskiy2021an} features into the 4th layer of the SiT-B/2, and the 8th layer of the SiT-XL/2, and mirror their hyperparameter setup.

\begin{table}[h]
    \centering
    \begin{subtable}{\linewidth}
        \centering
        \tablestyle{5pt}{1.1}
        \begin{tabular}{z{60}x{22}x{20}x{22}x{23}x{22}} 
        \toprule 
        \multirow{2}{*}{ } & \multicolumn{5}{c}{Metrics} \\ 
        \cmidrule{2-6}                  Model & FID $\downarrow$ & IS $\uparrow$ & sFID $\downarrow$ & Prec. $\uparrow$ & Rec. $\uparrow$ \\ 
        \midrule 
        REPA SiT-B/2                    & 27.33   & \phantom{0}61.60  & 11.70 & 0.57    & {\bf 0.64} \\ 
        \rowcolor{gray!25}
        % (\name{}) REPA SiT-B/2          & {\bf 20.52}   & \phantom{0}{\bf69.71}  & \phantom{0}{\bf 5.47} & {\bf 0.61}  & 0.63 \\ 
        + Using \name{}          & {\bf 20.52}   & \phantom{0}{\bf69.71}  & \phantom{0}{\bf 5.47} & {\bf 0.61}  & 0.63 \\ 
        % \cmidrule{1-5} 
        \midrule
        REPA SiT-XL/2                   & 11.14   & 115.83 & \phantom{0}8.25 & 0.67          & {\bf 0.65} \\ 
        \rowcolor{gray!25}
        % (\name{}) REPA SiT-XL/2         & {\bf 7.29} & {\bf 129.89} & \phantom{0}{\bf 4.93} & {\bf 0.71}    & 0.64 \\ 
        + Using \name{}         & {\bf 7.29} & {\bf 129.89} & \phantom{0}{\bf 4.93} & {\bf 0.71}    & 0.64 \\ 
        \bottomrule 
        \end{tabular}
        \caption{
        {\bf ImageNet-1k (256x256) Results with REPA.} Adding \cfm to REPA further improves SiT models across nearly all metrics, and by as much as 6.81 FID.
        }
        \label{tab:repa_imnet256}
    \end{subtable}

    \vspace{1em} % Space between the tables

    \begin{subtable}{\linewidth}
        \centering
        \tablestyle{5pt}{1.1}
        \begin{tabular}{z{60}x{22}x{20}x{22}x{23}x{22}} 
        \toprule 
        \multirow{2}{*}{ } & \multicolumn{5}{c}{Metrics} \\ 
        \cmidrule{2-6}        Model & FID$\downarrow$ & IS$\uparrow$ & sFID$\downarrow$ & Prec.$\uparrow$ & Rec.$\uparrow$ \\ 
        \midrule 
        REPA SiT-B/2                    & 31.90 & \phantom{0}56.96   & 13.78   & 0.67 & 0.62 \\ 
        \rowcolor{gray!25}
        % (\name{}) REPA SiT-B/2          & {\bf 24.48} & \phantom{0}{\bf 64.74  } & {\bf 5.89  } & {\bf 0.71} & 0.61\\ 
        + Using \name{}          & {\bf 24.48} & \phantom{0}{\bf 64.74  } & \phantom{0}{\bf 5.89  } & {\bf 0.71} & 0.61\\ 
        % \cmidrule{1-5} 
        REPA SiT-XL/2                   & 11.32 & 119.72    & 10.21 & 0.76  & {\bf 0.63 }\\
        \rowcolor{gray!25}
        % (\name{}) REPA SiT-XL/2         & {\bf 7.64}  & {\bf 131.50}    & {\bf 4.72}  & {\bf 0.79}  & 0.62 \\
        + Using \name{}         & {\bf 7.64}  & {\bf 131.50}    & \phantom{0}{\bf 4.72}  & {\bf 0.79}  & 0.62 \\
        \bottomrule 
        \end{tabular}
        \caption{
        {\bf ImageNet-1k (512x512) Results with REPA.} 
        \cfm is robust with REPA at large image resolutions, further improving performance across established metrics.
        }
        \label{tab:repa_imnet512}
    \end{subtable}

    \caption{
    REPA SiT~\cite{ma2024sit} results on ImageNet-1k ($256\times 256$; a) and ($512\times 512$; b).
    % $\uparrow$ indicates that higher values are better, with $\downarrow$ denoting the opposite.
    All models are trained for 400K iterations strictly following the procedure in~\citep{yu2024representation}, and set $\lambda=0.05$.
    We use the SDE Euler-Maruyama sampler with NFE=50 without classifier guidance for all our metrics.
    }
    \label{tab:repa_results}
\end{table}
\paragraph{Results.} We report results in Table~\ref{tab:repa_results}.
Similar to Section~\ref{sec:sit_vanilla}, \cfm substantially improves REPA models by as much as 6.81 FID, and consistently improves flow-matching with model scale. 
This highlights the versatility of the contrastive flow-matching objective as a broadly applicable criterion for diffusion model.

\begin{figure*}[!t]
\centering
    \includegraphics[width=0.95\textwidth]{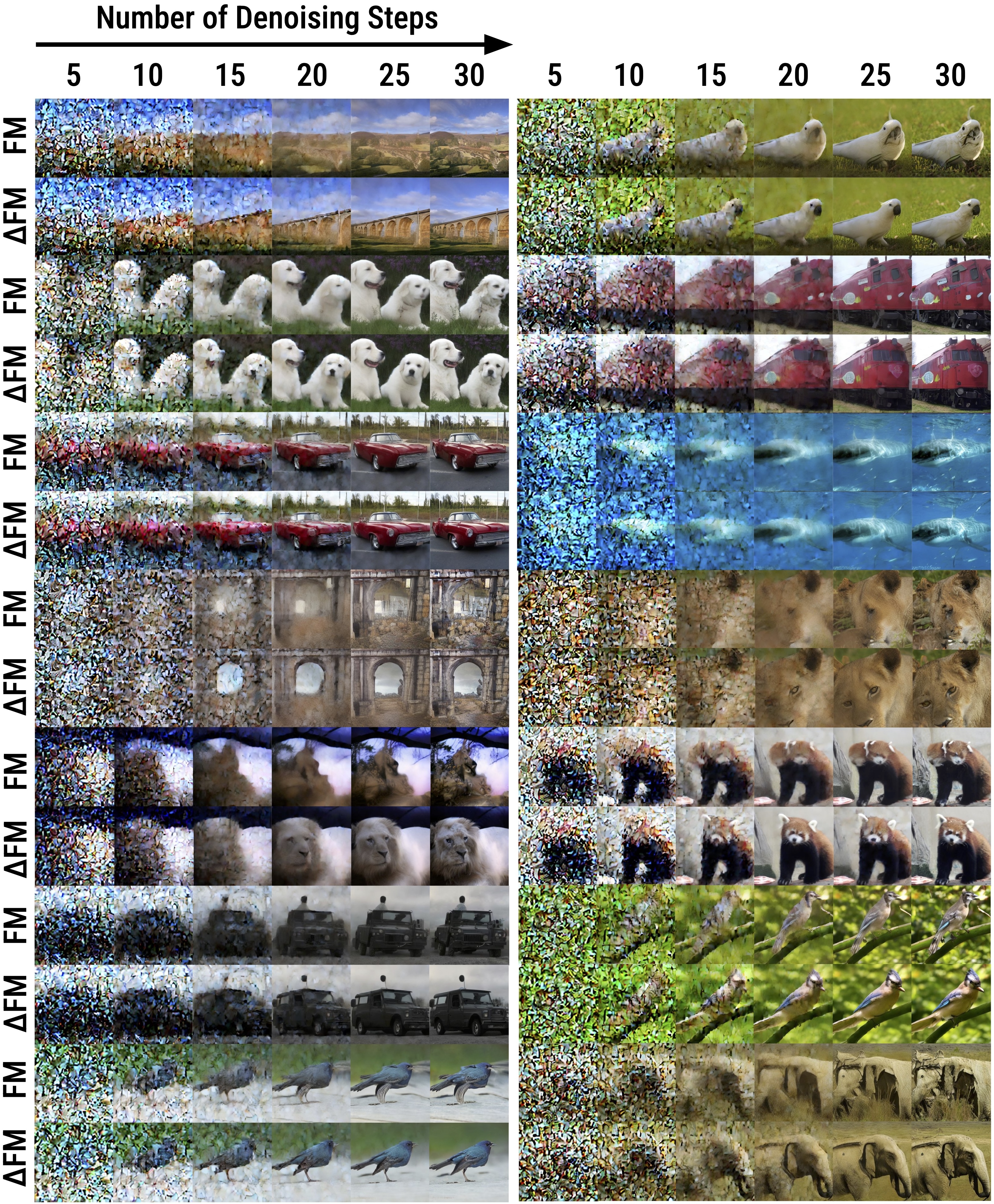}
\caption{
{\bf Contrastive flow-matching (\cfm) denoises significantly more efficiently than flow-matching.}
We visualize the expected final image estimated by a flow-model when denoised every 5 steps for trajectories of length 30 steps using the SDE Euler-Maruyama sampler and do not use classifier guidance. 
We compare the trajectories of a REPA SiT-XL/2~\cite{yu2024representation} trained on ImageNet-256~\cite{imagenet} for 400K steps with flow-matching (FM), and the same model trained with the contrastive flow-matching (\cfm) objective.
We show these trajectories in sets of pairs generated from the same noise sample during inference, with the flow-matching model above our \cfm version.
}
\label{fig:flow_trajectories}
\end{figure*}

\subsection{Extending to text-to-image generation}\label{sec:t2i}

\paragraph{Implementation Details.} We train models with the popular MMDiT~\cite{esser2024scaling} architecture from scratch on the CC3M dataset~\cite{sharma2018conceptual} for 400K iterations. For faster training, we pair each model with REPA, and follow the recommended training protocol of~\cite{yu2024representation}.
\begin{table}[h]
    \centering
    \tablestyle{5pt}{1.1}
    \begin{tabular}{z{60}x{60}x{60}}
        \toprule 
            \multirow{2}{*}{Metric} & \multicolumn{2}{c}{REPA-MMDiT}\\ 
            \cmidrule{2-3}                 & Flow-Matching & \name{} \\ 
            \midrule 
            FID$\downarrow$ & 24 & \textbf{19} \\
            \bottomrule 
            \end{tabular}
    \caption{\textbf{\name{} improves on CC3M 256$\times$256.} We use the SDE Euler-Maruyama sampler with NFE=50 without classifier-free guidance. 
    }
    \label{tab:cc3m_results}
\end{table}

\paragraph{Results.} Table~\ref{tab:cc3m_results} shows our results. \cfm improves over the flow matching baseline by $5$ FID, highlighting its seamless transferability to the broader text-to-image setting. We show qualitative results in Appendix~\ref{app:t2i}.

\subsection{CFG stacks with contrastive flow matching}\label{sec:cfg}

Contrastive flow matching offers advantages of Classifier-Free Guidance (CFG), without incurring additional computational costs during inference.
In this section, we demonstrate that when computational resources permit, combining \cfm with CFG can yield further performance enhancements.

\paragraph{Accounting for conflicts.} CFG and \cfm encourage flow matching model generations to be unique and identifiable, in different ways.
Specifically, \cfm trains models whose conditional flows are steered away from other arbitrary flows in the training data, regardless of generation state ($x_t$). 
In contrast, CFG steers generations away from the unconditional flow estimates based on $x_t$.
Thus, the signals from each may not always be aligned and naively coupling them may lead to conflicts and suboptimal generations.
Fortunately, we can quantify the amount of steerage \cfm applies on flow matching models by deriving the closed-form solution to
Eq.~\ref{eq:cfm}: $\min_{\theta} \mathcal{L}^{(\Delta\text{FM})}(\theta) = \sfrac{\left[(\min_{\theta} \mathcal{L}^{(\text{FM})}(\theta)) - \lambda \hat{T}\right]}{\left[1-\lambda\right]}$, where $\hat{T}$ is simply the \textit{mean} of all sample trajectories from the training set (please see App.~\ref{app:closed_form_derivation} for the full derivation). 
Thus, \cfm yields models which estimate flows away from the \textit{data-driven} unconditional trajectory, weighted by $\lambda$.
While optimizer and training dynamics cannot guarantee that all models trained with \cfm exactly decompose into these terms, $\hat{T}$ nevertheless approximates its effect on these models.
With $\hat{T}$, we can account for conflicts between \cfm and CFG by modifying the CFG equation to:
$\hat{\text{CFG}} = (1-\lambda) \left[ w v(x_t|y) + (1 - w)v(x_t|\emptyset) \right] + \lambda \tau$, where $w$ is the guidance scale, $\emptyset$ is the unconditional term and $\lambda$ is the same parameter used during \cfm training (Appendix~\ref{app:cfg_derivation} contains the full derivation).
Note that, we only apply $\hat{\text{CFG}}$ within the specified guidance interval $[\sigma_{\text{low}}, \sigma_{\text{high}}]$, and use our \textit{unchanged} \name{} model 
% when denoising 
outside this interval.

\begin{table}[t]
    \centering
    \tablestyle{5pt}{1.1}
    \begin{tabular}{z{55}x{15}x{15}x{15}x{20}x{20}x{20}} 
        \toprule 
            \multirow{2}{*}{Model} & \multicolumn{3}{c}{CFG Terms} & \multicolumn{3}{c}{Metric} \\ 
            \cmidrule{2-6}                 & $w$ & $\sigma_{\text{low}}$ & $\sigma_{\text{high}}$ & IS$\uparrow$ & FID$\downarrow$ & sFID$\downarrow$ \\ 
            \midrule 
            REPA SiT-XL/2                   & 1.75 & 0.0 & 0.75 & 280.33 & 2.09 & 5.55 \\
            \rowcolor{gray!25}
            + Using \name{}                 & 1.85 & 0.0 & 0.65 & {\bf 281.95} & {\bf 1.97} & {\bf 4.49} \\
            \bottomrule 
            \end{tabular}
    
    \caption{\textbf{ImageNet 256$\times$256 Results with CFG and NFE=50.} 
    ``$w$'' denotes the classifier-free guidance (CFG) weight, and $[\sigma_{\text{low}}, \sigma_{\text{high}}]$ is the time interval under which CFG is applied.
    We report the best results for each model after conducting a grid search over $w\in\{1.25, 1.75, 1.8, 1.85, 2.25\}$, $\sigma_{\text{low}}=0$ and $\sigma_{\text{high}}\in\{0.50, 0.65, 0.75, 1.0\}$.
    \cfm outperforms FM on all metrics.
    }
    \label{tab:cfg_results}
\end{table}

\paragraph{Results.} 
Table~\ref{tab:cfg_results} summarizes the results.
When paired with CFG, \cfm improves flow matching models across all metrics, demonstrating its efficacy in settings where computational costs are not a constraint. 

\paragraph{Additional Couplings.}
While we find that our proposed coupling strategy for \cfm and CFG works well for our setting, other suitable variations may also exist.
For instance, one may instead reduce conflicts by following the equation: $\Tilde{CFG} = (w+\lambda)v(x_t|y) - (1-w)v(x_t|\emptyset) - \lambda\hat{T}$, where $\lambda \text{, and } w$ are free hyperparameters. 
We leave such exploration to future work.

\subsection{Analyzing Contrastive Flow-Matching}\label{sec:analysis} 

\paragraph{Understanding the \cfm weight ($\lambda$).}
$\lambda$ directly controls how unique flows are across classes. 
Increasing $\lambda$ encourages every diffusion step to be fully discriminative, enabling models to encode distinct representations that integral to generating strong visual outputs at each trajectory step. 
However, setting it too high can lead to overly-separated flow trajectories, making it difficult to capture the class structure (Table~\ref{tab:varying_lambda}).
However, $\lambda$ values that are too low mirror the flow matching objective. 
Notably, we find that $\lambda=0.05$ is stable across all model and dataset settings, consistently achieving strong performance.
\begin{table}[t]
    \centering
    \tablestyle{5pt}{1.1}
    \begin{tabular}{z{30}>{\columncolor[gray]{.9}}x{20}x{20}x{20}x{20}x{20}x{20}} 
    \toprule 
    \multirow{2}{*}{Metric} & \multicolumn{5}{c}{\name{} $\lambda$ Values} \\ 
    \cmidrule{2-7}                  & $0.0$ & $0.001$ & $0.01$ & $0.05$ & $0.1$ & $0.15$ \\
    \midrule 
    IS$\uparrow$                    & 115.83 & 115.70 & 119.41 & {\bf 129.89} & 116.27 & 82.20 \\
    \cmidrule{2-7}
    FID$\downarrow$                 & 11.14 & 10.93 & 9.93 & {\bf 7.29} & 9.86 & 19.21 \\
    \bottomrule 
    \end{tabular}
    
    \caption{
    {\bf $\lambda=0.05$ is ideal.} 
    We show an ablation of the \cfm{} weight parameter $\lambda$. A too large $\lambda$ produces degenerate distributions that do not model class structure well. Too low $\lambda$ is essentially identical to flow-matching, with very little effect on training. $\lambda=0.05$ is best and we use this for all our experiments.
    }

\end{table}\label{tab:varying_lambda}

\noindent\textbf{Earlier class differentiation during denoising.} In Figure~\ref{fig:flow_trajectories}, we study flow trajectories of standard flow matching (FM) and flow matching with \cfm. To do this, we take partially denoised latents at various intermediate time steps along a trajectory with total length 30.
While initially both follow similar trajectories, they quickly diverge within the first several steps of the denoising process.
For instance, the model trained with \cfm produces more structurally coherent images earlier (around 15 to 20 steps in) than with FM. 
The iconic features of each class, such as slanted bridge surfaces (\Cref{fig:flow_trajectories} (top-left)), animal eyes (\Cref{fig:flow_trajectories} (upper-left and top-right), and train windows (\Cref{fig:flow_trajectories} (upper-right)), are more clearly visible early on during the diffusion process of the \cfm model. This enables \cfm to ultimately generate higher quality images at the final timestep.

\begin{table}[t]
\centering
\tablestyle{5pt}{1.1}
\begin{tabular}{z{60}x{40}x{22}x{22}x{22}} 
\toprule 
\multirow{2}{*}{ } & \multirow{2}{*}{ } & \multicolumn{3}{c}{Metrics} \\ 
\cmidrule{3-5}                  Model & Batch Size & FID $\downarrow$ & IS $\uparrow$ & sFID $\downarrow$ \\ 
\midrule 
REPA SiT-B/2                    & 256 & 42.28   & 38.04   & 11.35 \\ 
\rowcolor{gray!25}
% (\name{}) REPA SiT-B/2          & 256 & {\bf 33.39}   & {\bf 43.44}   & \bf{5.67} \\ 
+ Using \name{}          & 256 & {\bf 33.39}   & {\bf 43.44}   & \phantom{0}\bf{5.67} \\ 
\cmidrule{1-5} 
REPA SiT-B/2                    & 512 & 24.45 & 69.15 & 11.42 \\
\rowcolor{gray!25}
% (\name{}) REPA SiT-B/2          & 512 & {\bf 17.06} & {\bf 81.41} & {\bf 5.29} \\ 
+ Using \name{}          & 512 & {\bf 17.06} & {\bf 81.41} & \phantom{0}{\bf 5.29} \\ 
% \midrule 
\cmidrule{1-5}
REPA SiT-B/2                    & 1024 & 22.00 & 76.15 & 11.76 \\ 
\rowcolor{gray!25}
% (\name{}) REPA SiT-B/2          & 1024 & {\bf 15.23} & {\bf 88.53} & {\bf 5.20} \\ 
+ Using \name{}          & 1024 & {\bf 15.23} & {\bf 88.53} & \phantom{0}{\bf 5.20} \\ 
% \cmidrule{1-5} 
\midrule
REPA SiT-XL/2                   & 256 & 11.14      & 115.83       & 8.25 \\ 
\rowcolor{gray!25}
% (\name{}) REPA SiT-XL/2         & 256 & {\bf 7.29} & {\bf 129.89} & {\bf 4.93} \\
+ Using \name{}         & 256 & \phantom{0}{\bf 7.29} & {\bf 129.89} & {\bf 4.93} \\
\cmidrule{1-5}
REPA SiT-XL/2                   & 512 & 10.15 & 129.43 & 9.00 \\ 
\rowcolor{gray!25}
% (\name{}) REPA SiT-XL/2         & 512 & {\bf 6.36} & {\bf 146.17} & {\bf 5.42} \\ 
+ Using \name{}         & 512 & \phantom{0}{\bf 6.36} & {\bf 146.17} & {\bf 5.42} \\ 
\bottomrule \end{tabular}
\caption{
{\bf \name{} Scales with Batch Size.} 
We train all models for 400K iterations and strictly follow the protocol of \citep{yu2024representation}.
All metrics are measured with the SDE Euler-Maruyama sampler with NFE=50 and without classifier guidance.
We use $\lambda=0.05$ for all models trained with \name{} and do not change any other hyperparameters.
$\uparrow$ indicates that higher values are better, with $\downarrow$ denoting the opposite.
Improvement using \name{} evenly scales with batch-size, and even outperforms flow-matching models with \textit{half} the batch-size.
\vspace{1.3em}
}
\end{table}
\label{tab:res256_diff_bs}

\noindent\textbf{Effects of batch size on \cfm.} In Table~\ref{tab:res256_diff_bs}, we study the effects of batch size on our loss. It is well known that batch size has an important effect on contrastive style losses~\cite{caron2021emerging, chen2020simple, he2020momentum}
 that draw negatives within the batch. This can be understood as a sample diversity issue. If the batch size is larger than negative samples within the batch are more representative of the true distribution. In this table, we see a similar trend: larger batch sizes are important for maximizing the performance of \cfm across several model scales. We also maintain our improvements over the REPA baseline through all batch sizes and model scales.

\begin{figure}[t]
\centering
\begin{center}
    \includegraphics[width=1.0\linewidth]{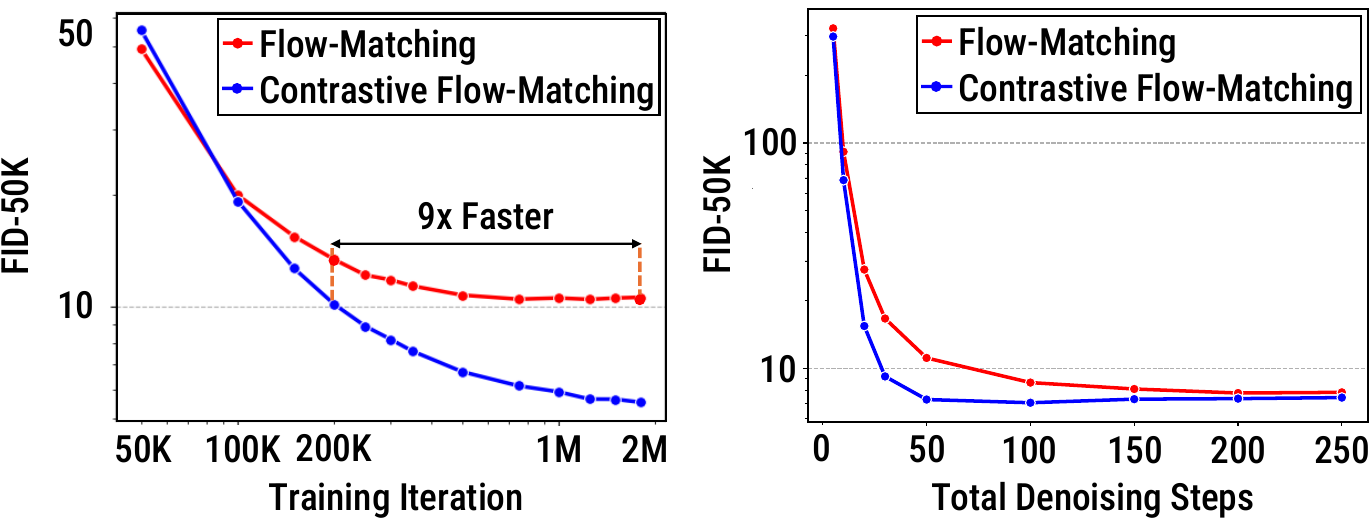}
\end{center}
\caption{
{\bf \cfm{} requires significantly fewer training iterations and inference-time denoising steps.} We plot FID-50k on ImageNet 256x256 with different numbers of training iterations and denoising steps. We see that \cfm outperforms the baseline with \textbf{9$\times$ fewer} training iterations and \textbf{5$\times$ reduction} in the number of inference-time denoising steps, indicating that \cfm is more efficient in both training and inference.
}
\label{fig:dne_steps}
\end{figure}

\noindent\textbf{Improved training and inference speed.} In Figure~\ref{fig:dne_steps} (left), we see the significant improvements in training speed from the \cfm objective. We reach the same performance (measured by FID-50k) as baseline with $9\times$ fewer training iterations. In Figure~\ref{fig:dne_steps} (right), we also demonstrate significant improvements at inference time. With our objective, we reach superior performance with only 50 denoising steps compared to the baseline with 250 denoising steps. This is a linear 5$\times$ improvement in training efficiency. Taken together, these results emphasize the important gains in computational efficiency achieved by our method. 

\section{Conclusion}

We introduced Contrastive Flow Matching (\cfm), a simple addition to the diffusion objective that enforces distinct, diverse flows during image generation. Quantitatively, \cfm results in improved image quality with far fewer denoising steps ($5\times$ faster) and significantly improved training speed ($9\times$ faster). Qualitatively, \cfm improves the structural coherence and global semantics for image generation. All of this is achieved with negligible extra compute per training iteration. Finally, we show that our improvements stack with the recently proposed Representation Alignment (REPA) loss, allowing for strong gains in image generation performance. Looking forward, \cfm shows the possibility that deviating from perfect distribution modeling in the diffusion objective might result in better image generation. 
\label{sec:conclusion}
{
    \small
    \bibliographystyle{ieeenat_fullname}
    \bibliography{main}
}
\clearpage
\appendix
\begin{figure*}[!ht]
\centering
\includegraphics[width=0.95\textwidth]{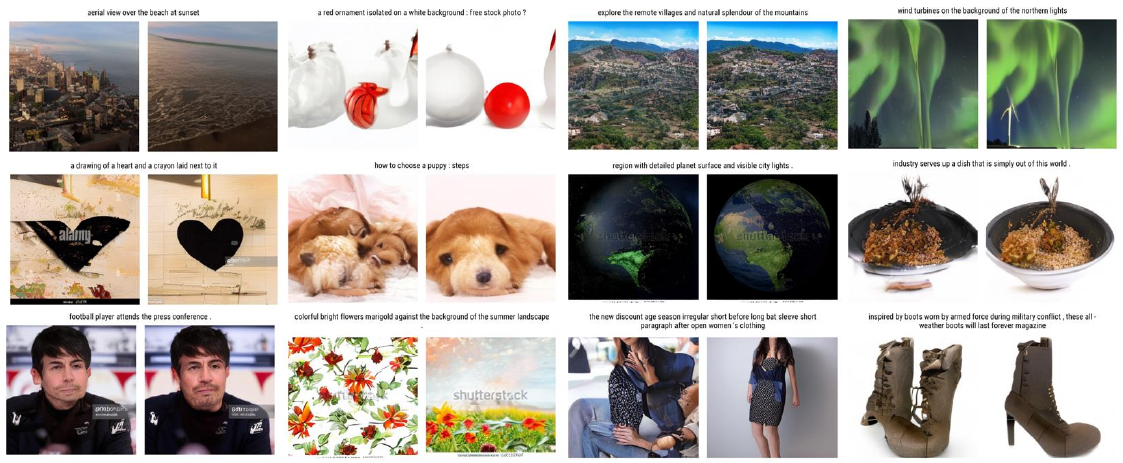}
\caption{
{\bf CC3M side-by-side generations between a REPA-MMDiT model trained with flow-matching (left) and \cfm (right).} 
Models are trained for 400K iterations using a batch-size of 256 and images are generated without classifier-free guidance and using NFE=50.
}
\label{fig:cc3m_pairs}
\end{figure*}

\section{Text-to-Image Qualitative Results}\label{app:t2i}
We visualize generations between our REPA-MMDiT models described in Section~\ref{sec:t2i} trained with flow-matching (FM) loss and with \cfm on CC3M with a batch size of 256 for 400K iterations in Figure~\ref{fig:cc3m_pairs}.
We plot images in pairs, with FM images on the left and \cfm images on the right, and show the respective caption for each pair above.
All images are generated without classifier-free guidance and using NFE=50, and are the same images used in Table~\ref{tab:cc3m_results}.

\section{Deriving Contrastive-Flow Matching Interference}\label{app:bias_derivation}
\subsection{Closed-form solution to Eq.~\ref{eq:cfm}}\label{app:closed_form_derivation}
We first re-introduce Eq.~\ref{eq:cfm} for convenience,
\begin{equation*}
\begin{split}
    \mathcal{L}^{(\cfm)}(\theta) = \mathrm{E}\left[
    \begin{aligned}
        & ||v_\theta(x_t, t, y) - (\dot{\alpha}_t \hat{x} + \dot{\sigma}_t \epsilon) ||^2 \\
        & -\lambda ||v_\theta(x_t, t, y) - (\dot{\alpha}_t \tilde{x} + \dot{\sigma}_t \tilde{\epsilon}) ||^2
    \end{aligned}
    \right]
\end{split}
\end{equation*}
Minimizing the expectation, expanding all norms and letting $v(\theta)=v(x_t,t,y)$, we can simplify the expectation to:
\begin{align}
    \begin{split}
        &= \min_\theta \mathrm{E}\left[
        \begin{aligned}
            (1-\lambda) v(\theta)^Tv(\theta) \\
            - 2v(\theta)^T\left[(\dot{\alpha}_t \hat{x} + \dot{\sigma}_t \epsilon) - \lambda (\dot{\alpha}_t \Tilde{x} + \dot{\sigma}_t \Tilde{\epsilon}) \right]\\
            + (\dot{\alpha}_t \hat{x} + \dot{\sigma}_t \epsilon)^T (\dot{\alpha}_t \hat{x} + \dot{\sigma}_t \epsilon) \\
            - \lambda (\dot{\alpha}_t \Tilde{x} + \dot{\sigma}_t \Tilde{\epsilon})^T (\dot{\alpha}_t \Tilde{x} + \dot{\sigma}_t \Tilde{\epsilon})
        \end{aligned}
        \right]
    \end{split} \\
    \begin{split}
        &= \min_\theta \mathrm{E}\left[
        \begin{aligned}
            (1-\lambda) v(\theta)^Tv(\theta) \\
            - 2v(\theta)^T\left[(\dot{\alpha}_t \hat{x} + \dot{\sigma}_t \epsilon) - \lambda (\dot{\alpha}_t \Tilde{x} + \dot{\sigma}_t \Tilde{\epsilon}) \right]
        \end{aligned}
        \right]
    \end{split} \\
    \begin{split}
        &\appropto \min_\theta \mathrm{E}\left[\left|\left|
        \begin{aligned}
            \sqrt{1-\lambda}v(\theta) \\
            - \frac{(\dot{\alpha}_t \hat{x} + \dot{\sigma}_t \epsilon) - \lambda (\dot{\alpha}_t \Tilde{x} + \dot{\sigma}_t \Tilde{\epsilon})}{\sqrt{1-\lambda}}
        \end{aligned}
        \right|\right|^2_2\right]
    \end{split}
\end{align}
Setting the gradient with respect to $v(\theta)$ to $0$,
\begin{align}
    \sqrt{1-\lambda}v(\theta)^* &= \mathrm{E}\left[\frac{(\dot{\alpha}_t \hat{x} + \dot{\sigma}_t \epsilon) - \lambda (\dot{\alpha}_t \Tilde{x} + \dot{\sigma}_t \Tilde{\epsilon})}{\sqrt{1-\lambda}}\right] \\
    v(\theta)^* &= \frac{ \mathrm{E}\left[ \dot{\alpha}_t \hat{x} + \dot{\sigma}_t \epsilon \right] - \lambda \mathrm{E}\left[ \dot{\alpha}_t \Tilde{x} + \dot{\sigma}_t \Tilde{\epsilon} \right] }{1-\lambda}
\end{align}
Finally, observe that $\mathrm{E}\left[ \dot{\alpha}_t \hat{x} + \dot{\sigma}_t \epsilon \right]$ is the solution to the flow-matching objective. Setting $\mathrm{E}\left[ \dot{\alpha}_t \Tilde{x} + \dot{\sigma}_t \Tilde{\epsilon} \right]=\hat{T}$ and observing that $x_t$ does not depend on $\hat{x}$ or $\hat{\epsilon}$ we obtain: 
\begin{align}
    \min_{\theta} \mathcal{L}^{(\Delta\text{FM})}(\theta) = \frac{\min_{\theta} \mathcal{L}^{(\text{FM})}(\theta) - \lambda \hat{T}}{1-\lambda}\label{eq:closed_form}
\end{align}

\subsection{Coupling with CFG}\label{app:cfg_derivation}
Classifier-free guidance (CFG) is originally defined over the flow-matching solution of $\min_{\theta} \mathcal{L}^{(\text{FM})}$.
Re-writing Eq.~\ref{eq:closed_form} and substituting it into the CFG equation, we obtain:
\begin{align}
    &CFG = wv^{(FM)}(x_t, t, y) + (1-w)v^{(FM)}(x_t, t, \emptyset) \\
    \begin{split}
    &=\left[
    \begin{aligned}
        w\left[(1-\lambda)v^{(\cfm)}(x_t, t, y) + \lambda \hat{T} \right] \\
        -(1-w) \left[ (1-\lambda) v^{(\cfm)}(x_t, t, \emptyset) + \lambda \hat{T}\right]
        \end{aligned}
        \right]
    \end{split} \\
    \begin{split}
        &=\left[\begin{aligned}
            (1-\lambda) \left[\begin{aligned} w v^{(\cfm)}(x_t, t, y) \\
            + (1-w)v^{(\cfm)}(x_t, t, \emptyset)
            \end{aligned}\right]
        \end{aligned} + \lambda \hat{T}
    \right]\end{split}\label{eq:cfg_cfm}
\end{align}
Letting $v(x_t|y)=v^{(\cfm)}(x_t, t, y)$ and $v(x_t|\emptyset)=v^{(\cfm)}(x_t, t, \emptyset)$, we obtain the Eq. from Section~\ref{sec:cfg}: $\hat{CFG} = (1-\lambda)\left[ wv(x_t|y) + (1-w)v(x_t|\emptyset)\right] + \lambda \hat{T}$.

% \subsection{Other CFG Couplings}
% While we find that coupling \cfm with CFG following Eq.~\ref{eq:cfg_cfm} works well for our settings, we note that other variations may exist.
% For instance, one may instead combine the two following the equation: 
% \begin{align}
%     \Tilde{CFG} = (w+\lambda)v(x_t|y) - (1-w)v(x_t|\emptyset) - \lambda\hat{T}
% \end{align}
% Where $\lambda, w$ are a free hyperparameters.
% We leave these explorations to future work.

\end{document}